# Commentary

## A commentary on "The Now-Or-Never Bottleneck: A Fundamental Constraint on Language", by Christiansen and Chater (2016)

*Ramon Ferrer-i-Cancho*[1]

In a recent article, Christiansen and Chater (2016) present a fundamental constraint on language, i.e. a now-or-never bottleneck that arises from our fleeting memory, and explore its implications, e.g., chunk-and-pass processing, outlining a framework that promises to unify different areas of research. Here we explore additional support for this constraint and suggest further connections from quantitative linguistics and information theory.

## (a) Further support for the now-or-never bottleneck

Memory limitations are at the core of the now-or-never bottleneck. In section 3 of their article, Christiansen and Chater (2016) review well-established facts from psychological experiments on memory constraints. Further support for this view comes from statistical research showing that, for instance, the probability that two syntactically related words are at a certain distance in a sentence decays exponentially with distance in sentences of the same length (Ferrer-i-Cancho 2004). This decay of probability might be mirroring an exponential decay of memory as a function of time. A power-law decay has been reported when mixing distances from sentences of different length (Liu 2007), but this result could be an artefact of mixing information from sentences of different length (Ferrer-i-Cancho and Liu 2014).

In section 6.1.2, the authors review the statistical evidence of one implication of the now-or-never bottleneck "*The prevalence of local linguistic relations*" (presented in section 1). Further confirmation of this prediction of the now-or-never bottleneck is provided by statistical analyses showing that

- About 50% of adjacent words are linked syntactically (Yuret 2006) and about 50% of dependencies involve adjacent words (Ferrer-i-Cancho 2004; Liu 2008). This might be reflecting the need to recode the input as it comes along (Christiansen and Chater 2016).
- Constraints of syntactic dependencies such as planarity (non-crossing dependencies) could be a side effect of pressure for linguistic relations to be local (Ferrer-i-Cancho and Gómez-Rodríguez 2016b and references therein).

The latter finding is of enormous theoretical importance: it frees grammar, the language faculty,… from the responsibility of the rather low frequency of crossing dependencies in real languages. Connecting the dots, one could conclude that the now-or-never bottleneck predicts a structural property of syntax. That complements connections between chunk-and-pass and a wide range of linguistic phenomena that the authors indicate at the end of section 6.1.2. From a theoretical perspective, dependency length minimization not only predicts that crossings dependencies should have a low frequency (Ferrer-i-Cancho 2014b, Ferrer-i-Cancho and Gómez-Rodríguez 2016b) but also the consistency of branching and its direction, and the fixed order of certain kinds of syntactic relationships, e.g. adjective-noun (Ferrer-i-Cancho 2015a-b). The now-or-never bottleneck is crucial for the development of a parsimonious theory of syntax (Ferrer-i-Cancho and Gómez-Rodríguez 2016a).

---

[1] Complexity and Quantitative Linguistics Lab. LARCA Research Group, Departament of Computer Science, Universitat Politècnica de Catalunya (UPC). Campus Nord, Edifici Omega, Jordi Girona Salgado 1-3. 08034 Barcelona, Catalonia (Spain). Phone: +34 934134028.
E-mail: rferrericancho@cs.upc.edu

Christiansen and Chater (2016) propose a unified approach across scales, from the sentence scale to the historical/biological scale passing through the individual scale (Fig 3 of their article). Interestingly, the hypothesis of a word order permutation ring that constrains word order evolution (Ferrer-i-Cancho 2015a) can be reinterpreted as another implication of the now-and-never bottleneck at the historical level. Under pressure for dependency length minimization, that ring predicts that the development of SVO order from SOV is more likely than the development of OVS from SOV, although both SVO and OVS are convenient from the perspective of dependency length minimization (Ferrer-i-Cancho 2015a), in full agreement with historical data (Gell-Mann and Ruhlen 2011).

## (b) The now-or-never bottleneck as a solution to puzzles

The now-or-never bottleneck and its implications can help us to understand various puzzles. As we mentioned above, the probability that a syntactic dependency involves words at a certain distance decays exponentially with distance but slows down for distances about 4-5 onwards in Czech (Ferrer-i-Cancho 2004; see Fig. 4 B and D). The point is: if the length of a syntactic dependency is a burden, why the decay of that probability slows down at some point? The paradox could be solved hypothesizing a chunk-and-pass strategy and multiple levels of representation to fight against memory limitations at long distances, supporting Christiansen and Chater's (2016) view. Interestingly, a distance of 4-5 could be related with the size of word chunks at some level. Suppose that dependents can go eihter to the left and to the right of their head. Then a crossover at distance $d$ translates into a chunk size of $2d+1$. Applying this theoretical argument, the crossover at distance 4-5 words yields chunk sizes of 9-10 words. In contrast, if dependents can go only at one side of their head (consistent branching), a crossover at distance $d$ translates into a chunk size of $d+1$ and then the crossover at distance 4-5 yields a chunk size of 5-6 words. The relationship between that crossover and chunk size should be investigated with the help of dependency treebanks.

Another puzzling fact comes from analyses of the growth of the sum of dependency lengths as a function of sentence length: this sum is below a random baseline (supporting dependency length minimization) but clearly above the theoretical minimum, the one that is obtained by solving the minimum linear arrangement problem (Ferrer-i-Cancho 2004, Ferrer-i-Cancho and Liu 2014). The question is: why are real languages not getting to that minimum? One answer is that dependency length minimization is in conflict with other principles or that pressure to minimize dependency lengths increases with the length of the sentence (Ferrer-i-Cancho 2014a). A more interesting possibility follows from Christiansen and Chater's now or never bottleneck: an online, incremental, chunk-and-pass language production would lead to suboptimal linear arrangements and the real sum of dependency lengths would reflect it. Real sentences are unlikely to be optimized following the batch processing that algorithms for solving the minimum linear arrangement problem imply (Chung 1984; Shiloach 1979).

*A commentary on "The now-or-never bottleneck: a fundamental constraint on language", by Christiansen and Chater (2016)*

## (c) Conflicts between implications of the now-or-never bottleneck

At the beginning of section 1, the authors enumerate the implications of the now-or-never bottleneck:

> 2. The prevalence of local linguistic relations
>
> …
>
> 4. The use of prediction in language interpretation and production.

Some aspects of these two implications are incompatible: the word order that maximizes locality is incompatible with the word order that maximizes prediction (Ferrer-i-Cancho 2014a). That conflict could be one of the reasons why the sum of dependency lengths of sentences does not get to the theoretical minimum (as explained in (b)). This kind of conflict is reminiscent of other conflicting constraints in language optimization at the level of the mapping of words into meanings (Ferrer-i-Cancho 2005).

Conflicts between implications are not a problem for the now-or-never bottleneck. They may simple illuminate the wide range of solutions adopted by language and why languages keep evolving.

## (d) Further insights from the information theory

In section 6.1.3, the Christiansen and Chater (2016) write "*The problem of encoding and decoding digital signals over an analog serial channel is well-studied in communication theory (Shannon, 1948)—and, interestingly, the solutions typically adopted look very different from those employed by natural language.*"

The authors are missing recent research based on information theory predicting solutions that resemble a lot those of natural language: duality of patterning (Plotkin and Nowak 2000; relevant for section 6.1.4), Zipf's law for word frequencies (Ferrer-i-Cancho 2016a, Prokopenko et al 2010, Ferrer-i-Cancho 2005), Zipf's law of abbreviation (Ferrer-i-Cancho et al 2015), Menzerath's law (Gustison et al 2016), Clark's principle of contrast (Ferrer-i-Cancho 2017), a vocabulary learning bias in children (Ferrer-i-Cancho 2017),…

Information theory can make even more general predictions on language and beyond. Power-law-like distributions such as Zipf's law for word frequencies can be regarded as manifestations of critical-like behaviour (Kello et al 2010). In general, the critical-like behaviour that many natural systems such as language exhibit can be explained using information theory: "*criticality turns out to be the evolutionary stable outcome of a community of individuals aimed at communicating with each other to create a collective entity*" (Hidalgo et al 2014).

As the authors say, "*the Now-or-Never bottleneck provides a constant pressure towards reduction and erosion across the different levels of linguistic representation, providing a possible explanation for why grammaticalization tends to be a largely unidirectional process*" but there are even stronger predictions that can be made from the now-or-never bottleneck. The authors argue that "*the brain must compress and recode linguistic input as rapidly as possible*" to deal with the "Now-or-Never" bottleneck. It is precisely a generalized principle of compression (the minimization of the mean energetic cost of units) that predicts three linguistics laws: Zipf's law for word frequencies (Ferrer-i-Cancho 2016a), Zipf's law of abbreviation (Ferrer-i-Cancho et al 2015) and Menzerath's law (Gustison et al 2016). Furthermore, that principle could be related to the principle of dependency length minimization: compression can help to reduce the actual distance between syntactically related words when measured in syllables or phonemes (Ferrer-i-Cancho 2015b). Thus, memory limitations could promote compression across levels and domains.

We hope that our comments stimulate further research in quantitative linguistics.

## Acknowledgements

We are grateful to N. Chater, M. Christiansen and C. Gómez-Rodríguez for helpful comments and stimulating discussions. This work was supported by the grants 2014SGR 890 (MACDA) from AGAUR (Generalitat de Catalunya) and also the APCOM project (TIN2014-57226-P) from MINECO (Ministerio de Economia y Competitividad).